\documentclass[a4paper]{svproc}

\usepackage[dvipsnames]{xcolor}
\usepackage{url}

\usepackage{url}
\usepackage{svg}

\usepackage[backend=biber,
            hyperref=true,
            url=false,
            isbn=false,
            doi=false,
            backref=false,
            style=ieee,
            natbib=true,
            mincitenames=1,
            maxcitenames=1,
            minbibnames=1, 
            maxbibnames=3, 
            citestyle=numeric-comp,
            sorting=nyt,
            block=none]{biblatex}
\usepackage{amsmath,amssymb}
\usepackage{graphicx}
\usepackage{soul}
\usepackage{adjustbox}
\usepackage{booktabs}
\usepackage{multirow}
\usepackage{verbatim}
\usepackage{xcolor} 
\usepackage{listings}
\usepackage{multicol}
\usepackage{diagbox}
\usepackage{subfig}
\usepackage{hyperref}
\usepackage{xspace}
\usepackage{textcomp}
\usepackage{gensymb}
\usepackage{marginnote}
\usepackage{tabularx}
\usepackage{wrapfig}
\usepackage{siunitx} 
\usepackage{caption}
\graphicspath{{figures/}}

\renewcommand{\remark}[3]{{\color{#2}[#1: #3]}}
\newcommand{\daniel}[1]{\remark{\textbf{Daniel}}{blue}{\textbf{#1}}}

\newcommand{\method}{MOTIF\xspace}

\begin{document}
\mainmatter
\title{The MOTIF Hand: A Robotic Hand for Multimodal Observations with Thermal, Inertial, and Force Sensors}
\titlerunning{MOTIF Hand}  
\author{Hanyang Zhou* \and Haozhe Lou* \and Wenhao Liu* \and Enyu Zhao \and \\ Yue Wang \and Daniel Seita \; {\footnotesize (* equal contribution)}
}

\authorrunning{Zhou, Lou, Liu, et al.} 
\tocauthor{Hanyang Zhou, Haozhe Lou, Wenhao Liu, Enyu Zhao, Yue Wang, Daniel Seita}
 
\institute{
Department of Computer Science, Viterbi School of Engineering\\
University of Southern California, Los Angeles CA 90089, USA,\\
\email{\{zhouhany,haozhelo,liuwenha,enyuzhao,yue.w,seita\}@usc.edu} 
}

\maketitle              

\vspace{-1em}
\begin{abstract}
Advancing dexterous manipulation with multi-fingered robotic hands requires rich sensory capabilities, while existing designs lack onboard thermal and torque sensing.  
In this work, we propose the \method hand, a novel multimodal and versatile robotic hand that extends the LEAP hand by integrating: (i) dense tactile information across the fingers (ii) a depth sensor, (iii) a thermal camera, (iv), IMU sensors, and (v) a visual sensor. 
The \method hand is designed to be relatively low-cost (under 4000 USD) and easily reproducible. 
We validate our hand design through experiments that leverage its multimodal sensing for two representative tasks. 
First, we integrate thermal sensing into 3D reconstruction to guide temperature-aware, safe grasping. Second, we show how our hand can distinguish objects with identical appearance but different masses—a capability beyond methods that use vision only. For more details, please check our website: \href{https://slurm-lab-usc.github.io/motif-hand/}{https://slurm-lab-usc.github.io/motif-hand}

\end{abstract}

\section{Introduction and Motivation}

Dexterous manipulation using multi-fingered hands has advanced rapidly, driven by innovations in both hardware design~\cite{shaw2023leaphand,romero2024eyesight} and algorithms for control~\cite{rubik_cube_2019,RobotSynesthesia2024}. Research in this area predominantly uses third-person or wrist cameras for perception, along with tactile sensors such as XELA uSkin sensors~\cite{guzey2023dexterity} or lower-cost alternatives like force-sensing resistors~\cite{touch-dexterity,RobotSynesthesia2024}. 
However, these sensing methods are insufficient in more complex real-world tasks that require reasoning about multiple sensing modalities. Onboard depth and force sensing could assist in tasks such as reaching into a bag or box, while thermal sensing is essential for cooking or safe human-robot interaction, yet current robotic hands lack these capabilities. 
While prior work has used thermal cameras to perceive liquids in robot manipulation~\cite{schenck2017visualclosed,schenck2017perceivingreasoningliquids}, no existing robotic hand integrates a thermal camera into a multi-fingered dexterous platform.

\begin{figure*}[t]
\centering
\includegraphics[width=1.0\textwidth]{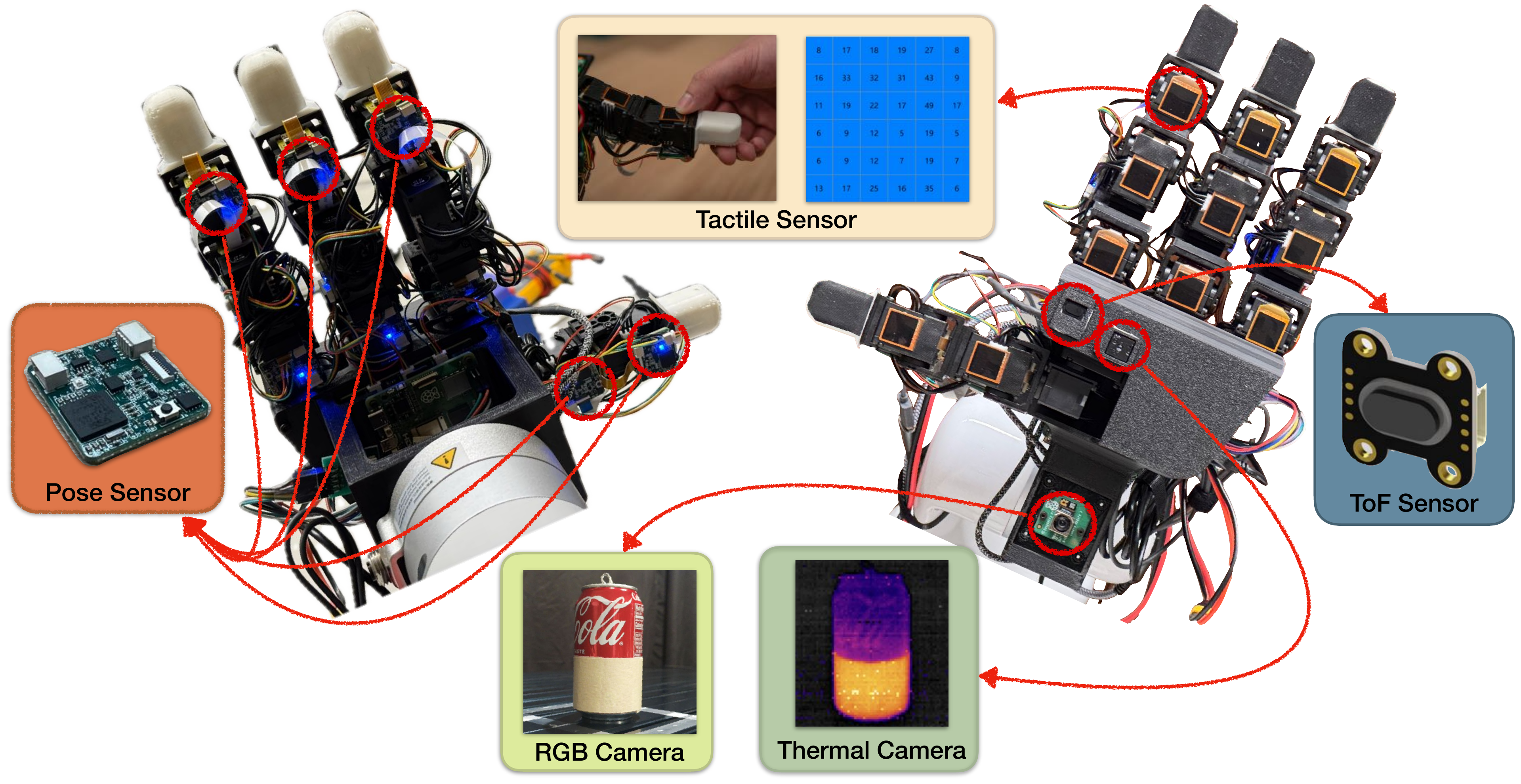}
\caption{
  The proposed \method hand, where we show the back (left) and front (right). We label the key components of the \method hand and show an example of data from the RGB and thermal cameras (using the can). See Section~\ref{sec:technical} for more details.
}
\vspace{-10pt}
\label{fig:pull}
\end{figure*}

To address these limitations, we introduce the \method hand (\textbf{M}ultimodal \textbf{O}bservation with \textbf{T}hermal, \textbf{I}nertial, and \textbf{F}orce sensors). This hand is built on the LEAP hand~\cite{shaw2023leaphand} and enhances its design with dense force and tactile sensing, along with onboard depth and thermal sensors (see Fig.~\ref{fig:pull}). By integrating these diverse sensing modalities, the \method hand provides comprehensive multimodal sensing capabilities that advance the current state of robotic manipulation research.
We will open-source our hand design and continually improve it to benefit the robotics community. 

\subsection{Related Work}

Prior work on dexterous manipulation with multi-fingered robotic hands falls into two broad categories: software-driven approaches and hardware innovations. 
Software-based approaches include learning-based algorithms for tasks such as in-hand reorientation using tactile-only feedback~\cite{qi2022hand,touch-dexterity} or visuo-tactile fusion~\cite{qi2023general,RobotSynesthesia2024}. Other works study tactile-driven manipulation using high-quality sensors on the hand~\cite{guzey2023dexterity}. 
Hardware innovations focus on improved tactile-based sensors.  
For example, modular magnetic sensors such as ReSkin~\cite{bhirangi2021reskin} and AnySkin~\cite{bhirangi2024anyskin} offer low-cost and replaceable sensing. 
Optical sensors such as the GelSight~\cite{yuan_gelsight_2017} and Digit 360~\cite{lambeta2024digitizingtouch} provide high-resolution touch feedback while pressure-based sensors convert mechanical force into electrical signals~\cite{huang20243dvitac}.  
In contrast to studies on individual sensors, we study full-hand designs, of which the LEAP hand~\cite{shaw2023leaphand} has become a widely-adopted platform in the robot manipulation community. 
Other recent hands include a five-finger variation of the LEAP hand, tested on dexterous cable manipulation~\cite{zhaole2025dexterouscablemanipulation} and the EyeSight hand~\cite{romero2024eyesight}, a 7-DoF humanoid hand that incorporates dense vision-based tactile sensors. 
However, no prior work has combined vision, depth, tactile, torque, and thermal sensing into a single hand, as we do with the \method hand. 


\section{Technical Approach: The \method Hand}
\label{sec:technical}
\vspace{-3pt}




We design our novel multisensory hand on top of the LEAP hand~\cite{shaw2023leaphand}. The \method hand includes thin film tactile sensors covering the fingers, a depth sensor, a thermal camera, and a high-resolution RGB camera on the wrist, as shown in Fig.~\ref{fig:pull}. We position accelerometers, gyroscopes, and magnetoresistive sensors at each finger joint and on the back of the hand to provide acceleration and orientation information.

For the finger pads, we use mature commercial tactile sensors for tactile perception input. The basic principle involves sensing force magnitude by measuring analog voltage signals in corresponding sampling areas of special materials with pressure-variable resistance values through array scanning. Our design provides sampling precision with six rows and six columns, each sub-unit having a resolution of 2.5mm. The trigger value for each sampling area is 20 g. 
While we include finger sensors, we do not use \emph{fingertip} sensors, as there are relatively mature and modular sensors such as the Digit 360~\cite{lambeta2024digitizingtouch} which we can use in a future version of the \method hand. 

The \method hand uses three processing tiers. 
On the \textbf{first-tier} processing unit at each finger joint, we install a motion tracking unit that combines a 3-axis gyroscope and a 3-axis accelerometer. Finger joint movement features short distances, high instantaneous acceleration, and high degrees of freedom. Therefore, using traditional six-axis pose sensors for integration produces significant drift. We adopt a nine-axis sensor solution widely used in drone applications, employing the BMM350 3-axis magnetic sensor. 
We also use high-shielding conductive adhesive tape to isolate electromagnetic interference from the servomotors to the magnetic sensors. 

Each first-tier processing unit on the finger joints connects via an RS485 serial bus and Modbus Protocol to the \textbf{second-tier} module on the palm responsible for data integration, encoding, and transmission. 
Similarly, we install nine-axis sensing units for pose calculation on this circuit board. To collect signals over a larger hand area, the \method hand uses four centrally symmetric Inertial Measurement Units (IMUs) for multi-point sampling fusion. Tactile data, gyroscope data, and acceleration data from the palm and from 11 finger joints encode and transmit through UART via USB-C port to the host computer.

For our \textbf{third-tier} data processing unit, we directly use a Raspberry Pi 5 which provides increased computational power for integrating multimodal sensory data. FLIR Lepton 3.5 collects infrared temperature data and a small time of flight (ToF) sensor provides real-time distance from the palm to contact surfaces. Infrared temperature data initially collects at $160 \times 120$ resolution and outputs at $1280 \times 960$ resolution after interpolation. 
For temperature sensing, we can either directly use the resulting temperature value matrices or assign colors based on relative temperatures to render a thermal image. We use the Raspberry Pi Camera Module 2 on the wrist to provide RGB image data at 30fps. 


\section{Experiments}
\vspace{-3pt}

We perform two sets of experiments to demonstrate the capabilities and versatility of the \method hand. First, in Section~\ref{ssec:temperature_grasping}, we evaluate how its thermal sensors enable safe grasping of objects with high-temperature regions. Then, in Section~\ref{ssec:flick}, we investigate how its force-based sensors can distinguish objects with different mass but the same shape through a simple ``flick'' action. 

\begin{figure*}[t]
\centering
\includegraphics[width=1.0\textwidth]{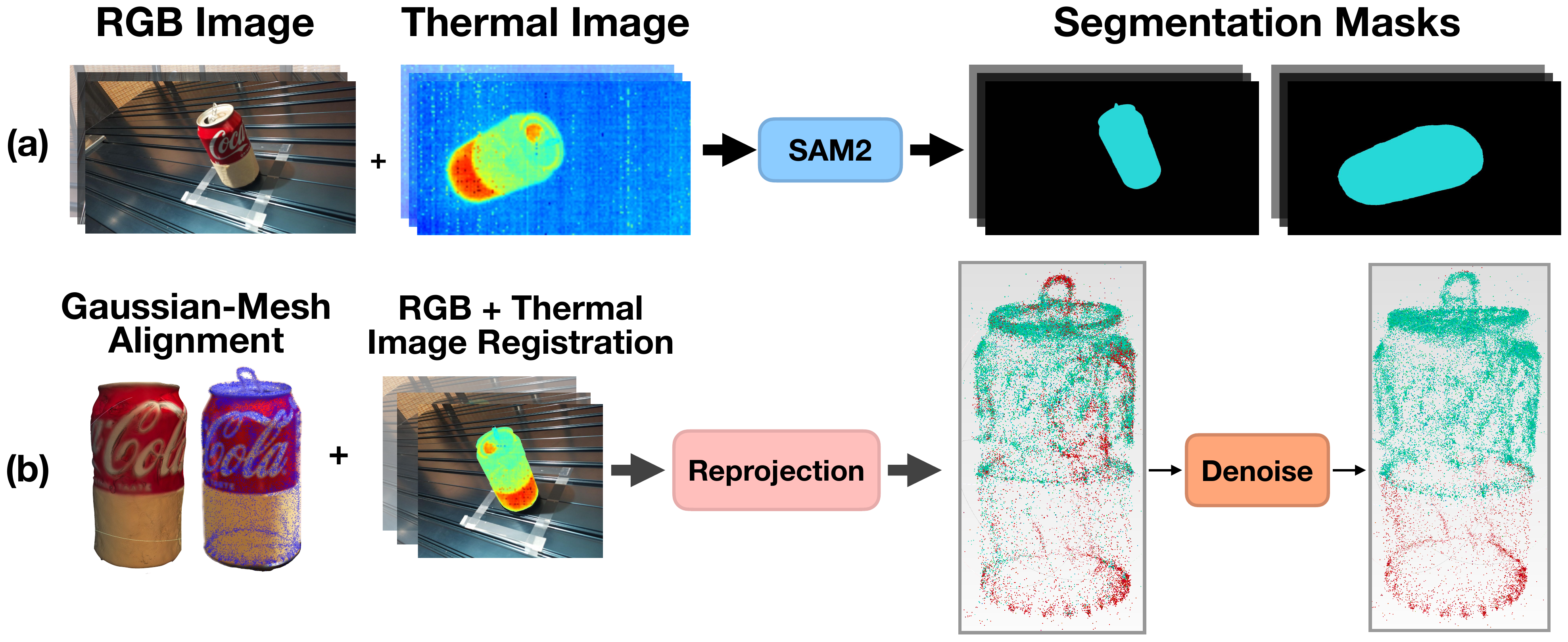}
\caption{
Data processing pipeline for thermal-based grasping. (a) First, we collect images from diverse viewpoints of the object and use SAM2~\cite{ravi2024sam2segmentimages} to extract the object mask. 
(b) We then reconstruct the 3D mesh and point cloud, perform thermal-RGB data alignment, and do reprojection. After denoising, we obtain our thermal affordance map which we pass to simulation for grasp pose optimization.  See Section~\ref{ssec:temperature_grasping} for details. 
}
\vspace{-10pt}
\label{fig:data_processing}
\end{figure*}

\vspace{-2pt}
\subsection{Temperature-Aware Object Modeling for Safe Grasping}
\label{ssec:temperature_grasping}

\begin{figure}[t]
    \centering
    \includegraphics[width=\textwidth]{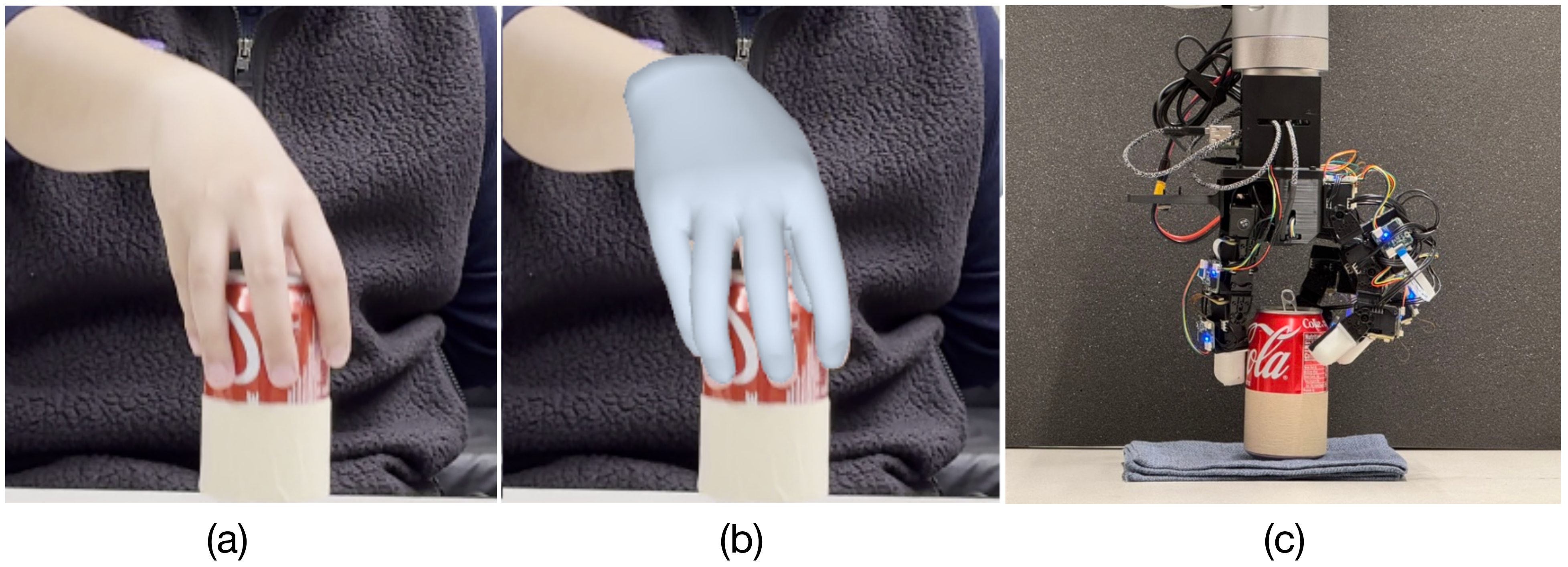}
 \caption{Hand pose and grasping demonstration pipeline. (a) A human hand performing one episode of demonstration which shows a natural grasping motion on a cylindrical object, avoiding the high-temperature region. (b) Pose extraction from the human demonstration. (c) Robotic hand trained to replicate the demonstrated grasping behavior, successfully learning to avoid the hazardous tape area like humans.}
 \label{fig:grasp}
 \vspace{-12pt}
\end{figure}

In this experiment, we present a representative example of how the \method hand’s multimodal sensing can inform grasping. We implement a ``Real2Sim'' pipeline that uses sensor data to build a geometric and thermal representation of an object for use in a physics simulator. It consists of reconstruction, alignment, denoising, and training a grasp policy. See Figures~\ref{fig:data_processing} and~\ref{fig:grasp} for an overview. 

\vspace{-12pt}
\subsubsection{Reconstruction}
This phase consists of three primary steps: camera pose estimation, dense reconstruction, and mesh extraction. We begin by moving the robot's end-effector to capture approximately 90 images of the target object from various viewpoints. Using Structure-from-Motion (SfM)~\cite{schonberger2016structure}, we obtain a sparse 3D point cloud along with estimated camera poses.
Subsequently, we apply Gaussian Splatting~\cite{huang20242d,kerbl20233d} to produce a dense reconstruction of the scene that captures fine geometric details and color variations. Utilizing the resulting dense Gaussian splats (3DGS) combined with normal constraints~\cite{ye2024stablenormal,lou2024robo}, we extract a refined mesh that accurately represents the object's surface geometry. This mesh can be exported to a physics simulation engine such as MuJoCo~\cite{mujoco}.

\vspace{-12pt}
\subsubsection{Alignment}
We next align the reconstructed simulation scene with the physical environment~\cite{lou2024robo}, with a focus on synchronizing thermal and RGB images to ensure scale and spatial consistency. 
First, we employ the Scale-Invariant Feature Transform (SIFT) algorithm~\cite{Lowe2004SIFT} to establish correspondences between thermal and RGB camera images. After identifying these correspondences, we apply a reprojection procedure to map values from the thermal images onto the reconstructed 3D points using the dense 3D Gaussian Splat representation. Let ${}^{w}T_c \in SE(3)$ denote the camera pose in the world coordinate system, and let $\mathbf{K} \in \mathbb{R}^{3\times 3}$ represent the camera intrinsic matrix. The transformation from pixel-level thermal coordinates $\mathbf{x_{2D}}$ to 3D coordinates $\mathbf{x_{3D}}$ is:
\begin{equation}
\mathbf{x_{3D}} = {}^{w}T_c^{-1} \left( d \cdot \mathbf{K}^{-1}
\begin{bmatrix}
x \\[6pt]
y \\[6pt]
1
\end{bmatrix} \right)
\end{equation}
where:
$\mathbf{x_{2D}} = [x, y]^T$ is the pixel coordinate in the thermal image and $d$ is its associated depth. 
Once we establish the correspondence between the point cloud and pixel coordinates, thermal values for each pixel can be assigned to the respective points in the 3D Gaussian Splats~\cite{zhou2024feature3dgssupercharging3d}.
The result is a static 3D thermal affordance map, which highlights thermal distribution across the object's surface. 

\vspace{-12pt}
\subsubsection{Denoising}

Noise from low-resolution thermal images is inevitable. As shown in Figure~\ref{fig:data_processing}, the Coke can's top hole introduces spurious hot signals during reprojection. To mitigate this, we assume a planar boundary separates the liquid versus air regions inside the object. We identify this boundary by analyzing thermal gradients through cross-sectional slicing. We compute ``thermal values'' using a weighted sum of RGB values, with a higher coefficient for red values since high-temperature pixels have stronger red responses. 

We then search for maximum negative transitions from positive thermal values (undersurface, warm) to negative values (above-surface, cool). The priority goes to transitions where the upper layer exceeds a threshold, the lower layer falls below another threshold, and the jump exceeds 30 units. Otherwise, we select the largest negative transition among adjacent layers. This approach relies on expecting significant temperature gradients on water surfaces.
Using the identified boundary, we analyze point color values in both directions to determine dominant color patterns. Anomalous points, such as hot spots above water and cold spots below, are treated as noise and replaced with the dominant colors for each layer. This removes artifacts while preserving legitimate thermal features.

\vspace{-12pt}
\subsubsection{Grasping Policy Conditioned on Thermal Observations}
Given the 3D thermal affordance map and the Real2Sim alignment, we develop a grasping policy explicitly conditioned on thermal observations. The policy aims to avoid grasping points that could lead to interactions with thermally sensitive or potentially hazardous areas.
We utilize imitation learning based on human demonstrations~\cite{dexmv} and employ MuJoCo simulation~\cite{mujoco} to infer contact regions between the robotic grasping poses and the object. Specifically, we extract 5000 demonstration frames from 12 episodes, retarget them to our MOTIF hand, and subsequently use these frames to train the policy~\cite{qin2023anyteleop,pavlakos2024reconstructing}.
If the robotic hand encounters a high-temperature region on the object, the policy adjusts by shifting the robot's end-effector grasping pose away from the hazardous region.

Grasp poses are filtered based on proximity to high-temperature zones on the object’s thermal map. Only grasping configurations that avoid these heated regions are selected for execution. Finally, validated grasps are performed in the real-world robotic manipulation setup, ensuring safe and efficient interactions with the object based on its thermal characteristics.

\subsection{Multimodal Sensor Object Mass Classification via Flicks}
\label{ssec:flick}

Next, we conduct an experiment to test how the \method hand's multimodal accelerometer, gyroscope, and magnetometer sensors can help distinguish objects of different masses via a simple fingertip ``flick'' action.

\vspace{-9pt}
\subsubsection{Experiment}
For humans, a simple way to distinguish the weights of different objects is to perform a short-distance movement, such as gently flicking them with the fingertips. 
Thus, we design an experiment where the hand uses its index fingertip to strike three U-shaped objects of the same material but of different weights, applying the same torque. As shown in Fig.~\ref{fig:flick}(a), the lightest red object (82 g) has the highest displacement, the medium-weight blue object (125 g) has a slightly lower displacement than the first one, and the heaviest purple object (219 g) has the shortest displacement.

\vspace{-9pt}
\subsubsection{Data Collection}
\begin{figure}[t]
    \centering
    \includegraphics[width=1.00\textwidth]{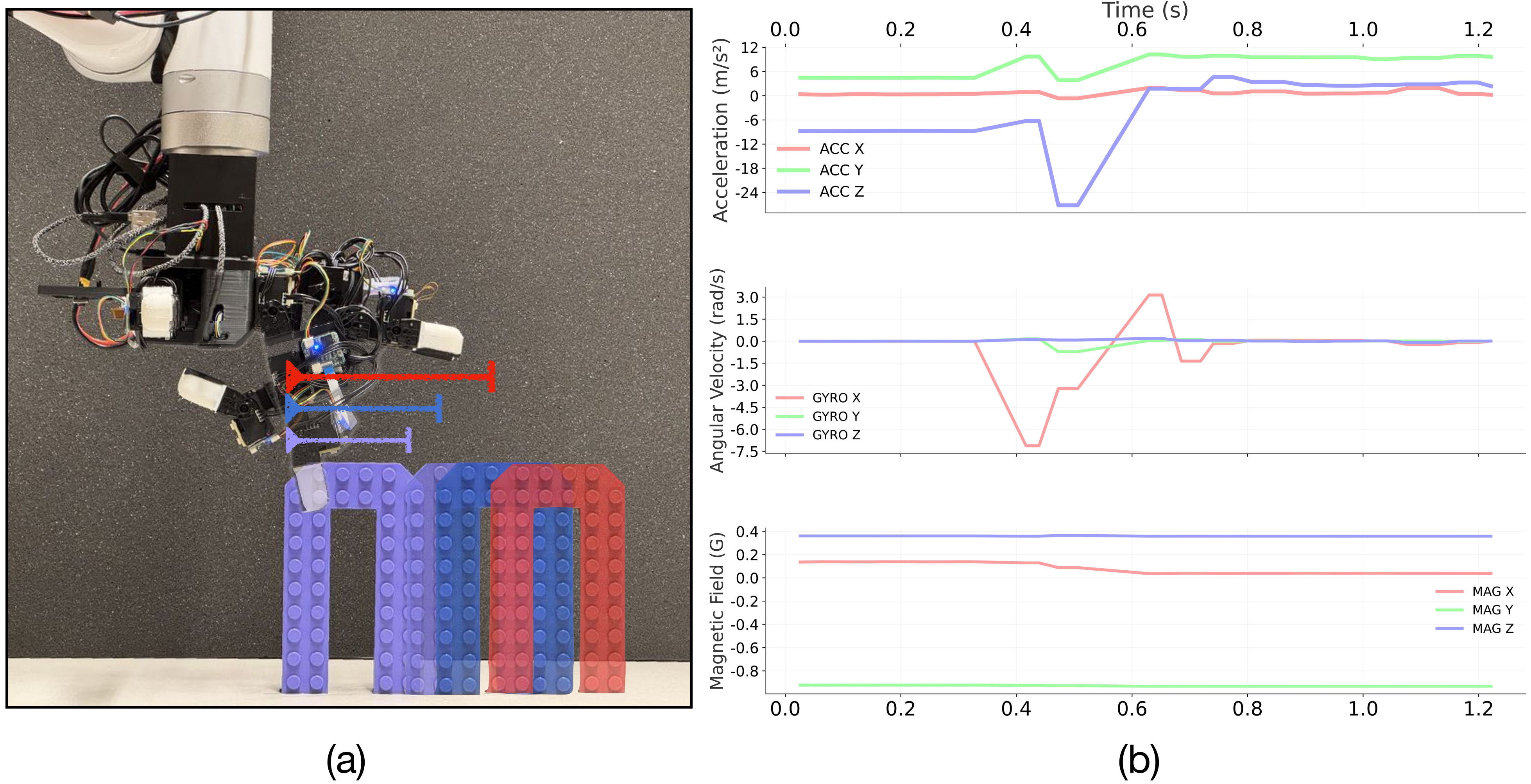}
 \caption{Fingertip flicking experiment setup and sensor data acquisition. (a) A visualization of how the \method hand's flick action affects the configuration of three U-shaped objects of different masses: purple (219 g), blue (125 g), and red (82 g). (b) Representative sensor data from the hand's accelerometer, gyroscope, and magnetometer during a single flicking trial, captured from 0.25 seconds before to 1 seconds after the flick action.}
 \label{fig:flick}
 \vspace{-10pt}
\end{figure}

The act of flicking objects with varying physical properties results in measurably different hand motion signatures, as recorded by sensors (accelerometer, gyroscope, and magnetometer) across their three-axis measurement frameworks.  Here, we focus on the measurements from the first board near the fingertip which is at the point of contact. To enable real-time data acquisition, the system transmits data at an update interval of 2 ms, with a Baud Rate of 115200 Bits/s. Data recording starts 0.25 seconds before performing the flick, and continues for another 1.0 seconds (see Fig.~\ref{fig:flick}(b)). We conducted 50 experiments on each object.

\vspace{-9pt}
\subsubsection{Classification Results and Analysis}
To validate the discriminative capability of the collected IMU sensor data, we performed Linear Discriminant Analysis (LDA)~\cite{Bishop2006PRML} on the sensor features from the flicking experiments. From the raw time-series data collected during each flicking motion, we extracted 42 statistical features to capture the essential characteristics. These features were derived from the sensors as follows: for the axis (X, Y, Z) of the accelerometer, gyroscope, and magnetometer, we computed four statistical measures: minimum, maximum, mean, and standard deviation. Additionally, we calculated the range (max-min) and magnitude mean (average of the three-axis combined magnitude) for each sensor type. This provides 14 features per sensor (3 axes × 4 statistics + 1 range + 1 magnitude), totaling 42 features across all three sensors (14 × 3 = 42). 

\begin{figure}[t]
    \centering
    \includegraphics[width=\textwidth]{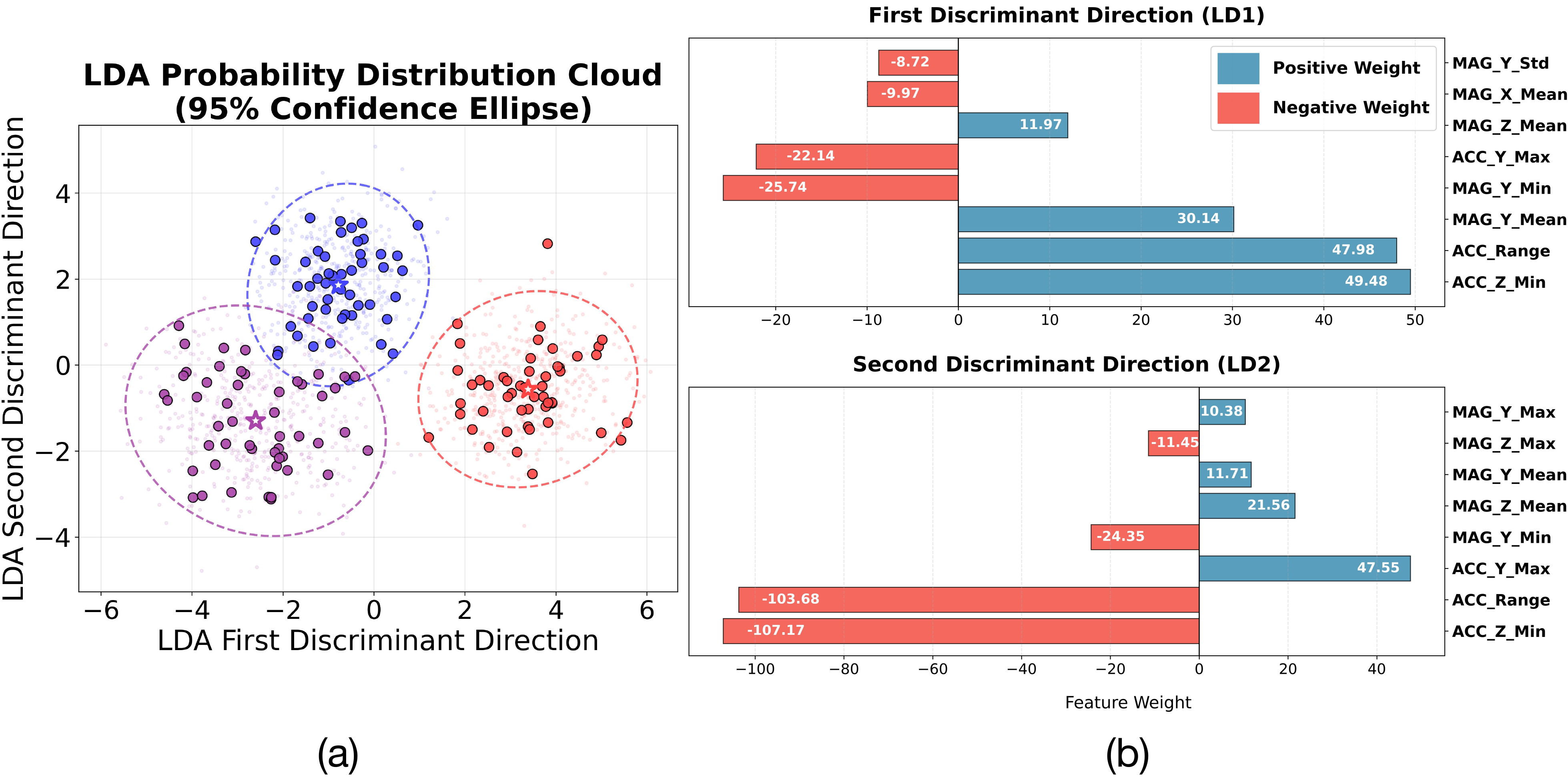}
 \caption{LDA analysis of multimodal sensor data from fingertip flicking experiments. (a) Two-dimensional LDA feature space showing separation of three object classes (red: 82 g, blue: 125 g, purple: 219 g) within 95\% confidence ellipses. (b) Feature contribution weights for the first and second discriminant directions, highlighting the dominant role of acceleration-based features in object mass classification.}
 \label{fig:flickanalysis}
 \vspace{-10pt}
\end{figure}

The results, shown in Fig.~\ref{fig:flickanalysis}, demonstrate strong separability among the three classes. The first discriminant direction (LD1) explains 77.5\% of the inter-class variance, while the second discriminant direction (LD2) accounts for the remaining 22.5\%, together capturing 100\% of the discriminative information. As shown in Fig.~\ref{fig:flickanalysis}(a), the three object categories form distinct clusters in the LDA feature space, with minimal overlap between their 95\% confidence ellipses. 

The feature contribution analysis in Fig.~\ref{fig:flickanalysis}(b) reveals physically meaningful patterns that align with the biomechanics of the fingertip flicking motion. The primary discriminant direction is mainly influenced by acceleration-related features, particularly \texttt{ACC\_Z\_Min} (49.48) and \texttt{ACC\_Range} (47.98). This is bio-mechanically reasonable because during the flick motion, the finger initially accelerates downward (negative Z-direction) to contact the object, then rapidly decelerates upon impact with an object of different weight. Heavier objects require greater contact force and create stronger resistance, resulting in more pronounced negative Z-acceleration values and larger acceleration ranges. The significant contribution of \texttt{MAG\_Y\_Mean} (30.14) reflects the lateral finger movement during the flicking gesture, as different object masses affect the finger's trajectory and stabilization requirements.

The second discriminant direction shows \texttt{ACC\_Z\_Min} (-107.17) and \texttt{ACC\_Range} (-103.68) with strong negative contributions, while \texttt{ACC\_Y\_Max} (47.55) provides positive discrimination. This pattern captures the finger's recovery motion after contact—lighter objects allow for quicker finger retraction with higher upward acceleration (Y-direction), while heavier objects result in more prolonged contact and different rebound dynamics. 

\vspace{-3pt}
\section{Experimental Insights}
\vspace{-3pt}

Multiple sensory signals, such as tactile, force/torque,  and temperature, play critical roles in human skin sensing. Tactile sensing helps reveal an object’s geometry and contact properties. Force sensing indicates the strength of interactions, enabling better control. Temperature sensing expands perception beyond touch, allowing us to infer thermal affordances and avoid damage from extreme temperatures. 
IMU sensors can support dynamic or closed-loop contact-rich tasks, such as in-hand manipulation and reorientation~\cite{qi2025simplecomplexskillscase}. 
We are investigating the essential combination of sensory modalities necessary for robotic manipulation tasks to ensure generalization across common indoor environments, such as kitchens. 
Therefore, we integrate visual, tactile, and proprioceptive sensors into our \method hand. Compared to relying solely on vision, our multimodal system enhances a robot's ability to detect, grasp, and manipulate objects, enabling more effective and safer interactions within complex real-world settings.

\vspace{-12pt}
\subsubsection{Real-World Applications.}

Manipulation in cooking environments is challenging due to the diverse and delicate foods, tools, and surroundings. The presence of hot fire, metal or wood cooking devices, and smooth-surfaced objects make it difficult to deploy traditional vision-based manipulation policies. 
However, leveraging the high-frequency in-hand sensors of the \method hand offer potential to capture underlying physical properties that are hard to determine from visual data alone. 
Beyond cooking, our hand can be valuable for industrial tasks such as welding and screw-tightening tasks in factories. A robotic welding system requires knowledge of when its welding gun tip is hot to complete the task. 
Additionally, screw-tightening demands high-precision positioning and real-time feedback of finger force and torque to achieve stable and accurate operations. 
By using the \method hand, we aim to make progress towards these tasks. 

\vspace{-12pt}
\subsubsection{Future Developments.}
More broadly, we envision that our \method hand has potential to be the first of many future iterations for improved multimodal robotic hands. This mirrors prior progress in robotic hands. After the original LEAP hand~\cite{shaw2023leaphand} was developed, 
researchers extended its design, such as by adding a fifth finger~\cite{zhaole2025dexterouscablemanipulation}. Furthermore, the original research team recently proposed the second generation of the LEAP hand~\cite{shaw2024leap}, which adds a soft exterior skin to create an anthropomorphic hand for robot learning. We plan to continually improve and expand our \method hand with more features. For example, the fingertips do not have sensors, so we will integrate the Digit 360 sensors~\cite{lambeta2024digitizingtouch} on them, pending availability. 
We hope that this inspires future work in developing methods for multimodal sensory robotic dexterous hands for complex tasks.

\vspace{-12pt}
\appto{\bibsetup}{\sloppy}
{
\begingroup
\renewcommand{\bibfont}{\small}
\printbibliography
\endgroup
}

\end{document}